\documentclass[letterpaper, 10 pt, conference]{ieeeconf}  

\usepackage{FG2024}
\usepackage{epsfig}
\usepackage{graphicx}
\graphicspath{{./images/}}
\usepackage{amsmath}
\usepackage{amssymb}
\usepackage{caption}
\usepackage[T1]{fontenc}
\usepackage{multirow}
\usepackage{pifont}
\usepackage{multirow}
\usepackage{colortbl}

\FGfinalcopy 

\IEEEoverridecommandlockouts                              

\def\FGPaperID{236} 

\title{\LARGE \bf
Cross-Block Fine-Grained Semantic Cascade for Skeleton-Based \\ Sports Action Recognition
}

\author{\parbox{16cm}{\centering
   {\large Zhendong Liu$^1$,  Haifeng Xia$^1$,  Tong Guo$^1$, Libo Sun$^2$, Ming Shao$^3$ and Siyu Xia$^1$}\\
    {\normalsize
    $^1$ School of Automation, Southeast University, Nanjing, China \\
    $^2$ School of Instrument Science and Engineering, Southeast University, Nanjing, China \\
    $^3$ Department of Computer and Information Science, University of Massachusetts Dartmouth, USA}}
    \thanks{This work was supported in part by the National Science Foundation under Grant No. 2144772 and in part by the Key R\&D Program of Jiangsu Province under Grant BE2023010-3. }
}

\usepackage{fancyhdr}
\begin{document}

\ifFGfinal
\thispagestyle{empty}
\pagestyle{empty}
\else
\author{\parbox{16cm}{\centering
    {\large Zhendong Liu$^1$,  Haifeng Xia$^1$,  Tong Guo$^1$, Libo Sun$^1$, Ming Shao$^1$ and Siyu Xia$^2$}\\
    {\normalsize
    $^1$ Faculty of Electrical Engineering, Mathematics and Computer Science, University of Twente, Enschede, The Netherlands\\
    $^2$ Department of Electrical Engineering, Wright State University, Dayton, USA}}
    \thanks{This work was not supported by any organization}
}
\pagestyle{empty}
\fi
\maketitle

\thispagestyle{fancy}

\begin{abstract}

Human action video recognition has recently attracted more attention in applications such as video security and sports posture correction. Popular solutions, including graph convolutional networks (GCNs) that model the human skeleton as a spatiotemporal graph, have proven very effective. GCNs-based methods with stacked blocks usually utilize top-layer semantics for classification/annotation purposes. Although the global features learned through the procedure are suitable for the general classification, they have difficulty capturing fine-grained action change across adjacent frames -- decisive factors in sports actions. In this paper, we propose a novel ``Cross-block Fine-grained Semantic Cascade (CFSC)'' module to overcome this challenge. In summary, the proposed CFSC progressively integrates shallow visual knowledge into high-level blocks to allow networks to focus on action details. In particular, the CFSC module utilizes the GCN feature maps produced at different levels, as well as aggregated features from proceeding levels to consolidate fine-grained features. In addition, a dedicated temporal convolution is applied at each level to learn short-term temporal features, which will be carried over from shallow to deep layers to maximize the leverage of low-level details. This cross-block feature aggregation methodology, capable of mitigating the loss of fine-grained information, has resulted in improved performance. Last, FD-7, a new action recognition dataset for fencing sports, was collected and will be made publicly available. Experimental results and empirical analysis on public benchmarks (FSD-10) and self-collected (FD-7) demonstrate the advantage of our CFSC module on learning discriminative patterns for action classification over others.


\end{abstract}

\section{Introduction}
In recent years, human action recognition has attracted more research interest in the computer vision community due to considerable application demands, especially for video surveillance and virtual reality games \cite{xia2023few}. Applications in these fields must recognize several continuous yet well-differentiable actions, such as walking, running, and jumping. These actions with notable differences have been the research interests of most existing approaches, which can be further divided into two branches: RGB-based and skeleton-based.

RGB-based action classification methods \cite{a7, a9, a10, a11, a14} extract salient motion information between frames by introducing high-cost computational optical flow or using different sampling frequencies. These works are highly dependent on task-irrelevant content, such as background, and are susceptible to environmental noise disturbances. Skeleton-based methods \cite{a23, a25, a27, a28, a30, a33} first employ pose estimation models \cite{a1, a2, a3} to extract 2D or 3D coordinates from human joints. Subsequently, the extracted coordinates are fed into GCN where they interact and integrate spatial information among distinct joints. This manner focuses on leveraging the underlying topological structure. Additionally, temporal edges are incorporated across consecutive frames to consolidate temporal features. 
\begin{figure}[t]
    \centerline{\includegraphics[width=80mm, height=80mm]{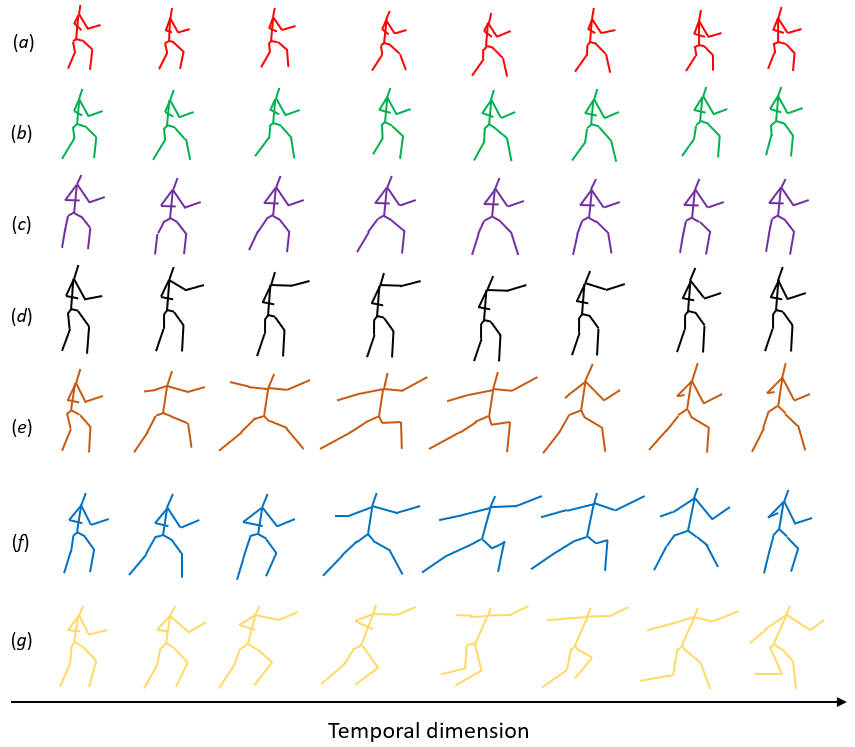}}
    \caption{Examples of Skeleton sequences from our self-built fencing dataset FD-7. (a) step forward, (b) two-step forward, (c) step backward, (d) thrust in place, (e) lunge in place, (f) step forward lunge, and (g) sprint.}
    \label{fig1}
\end{figure}
Subsequent research endeavors have introduced refined GCNs to effectively capture conspicuous motion features. For instance, \cite{a24} utilizes an adaptive GCN to increase flexibility in graph construction, and \cite{a33} learns a unique adjacency matrix for each channel of the feature map, improving the ability to aggregate information. Furthermore, \cite{a38} introduces a hierarchically decomposed graph to effectively identify important long-range connections between joint sets at the same hierarchical level. GCN-based methods suggest higher robustness and lower computational requirements, emerging as the predominant approach to address actions with substantial variations.

As a notable subdomain within action classification tasks, fine-grained classification of sports actions has garnered attention due to its significant effects on sports training assistance, sports competition analysis, and sports medicine rehabilitation. Examples of skleton sequences of sport actions can be found in Fig.~\ref{fig1}. Compared to other actions, sports actions are characterized by the following aspects. (1) They exhibit higher velocity and encompass a greater volume of motion information. For example, within the realm of fencing, the velocity at the tip of the sword can attain an impressive maximum of 150 meters per second, while in the context of figure skating, athletes can achieve remarkable rotational speeds of up to six revolutions per second following a jump. (2) They display minimal inter-class variance, where the visual distinctions among subordinate classes are often subtle due to their shared super-category. Within the realm of figure skating, jumps can be classified into diverse categories, contingent upon the skater's utilization of takeoff and landing edges, as well as the number of rotations executed midair. Consequently, this leads to a plethora of remarkably similar combinations. (3) They also demonstrate significant intra-class variance, as objects within the same category typically exhibit notable variations in poses and viewpoints. However, GCN-based methods are mostly focused on constructing generalized topology graphs and aggregating features within blocks. Thus, this manner lacks the utilization of low-level detail features and cross-block feature fusion.

Contemporary studies \cite{a48, a50, a52, a54} have provided compelling evidence that convolutional layers at different depths encompass information across diverse granularities. For instance, in the context of fine-grained image classification, when an image is fed into a convolutional neural network, shallow modules excel at capturing low-level features like edges and textures. As we progress towards intermediate modules, a deeper understanding of the image semantics gradually emerges and facilitates the extraction of larger-scale structural features. Finally, the deep modules consolidate the highest-level information, representing a comprehensive amalgamation of object parts and overall contours. Despite the semantic richness of high-level features, they inevitably sacrifice small-scale discriminative information. 
These missing fine-grained details carry significant importance in both image-based fine-grained classification and video-based fine-grained action recognition tasks.

To solve the fine-grained problem in action recognition, a plug-and-play Cross-block Fine-grained Semantic Cascade (CFSC) module is proposed. It captures low-level detail features that are gradually neglected during forward propagation to improve the performance of sports action classification.
Concretely, our approach involves the initial extraction of feature maps from distinct depth blocks at various levels. These feature maps serve as the foundation for capturing short-term temporal dependencies through the application of multiple small-kernel temporal convolutions. Subsequently, a process of aggregating feature maps from shallow to deep is undertaken to effectively fuse multi-granular information. Finally, discriminative representations that encompass rich low-level details are leveraged for classification. Our module is trained in conjunction with the backbone and is compatible with most GCN-based models. To substantiate the efficacy of our approach, we construct a fencing dataset, FD-7, encompassing seven distinct categories of common fencing actions. Specifically, (a) step forward, (b) two-step forward, and (c) step backward represent fencing footwork, while (d) thrust in place, (e) lunge in place, (f) step forward lunge, and (g) sprint represent fencing attacks, as shown in Fig.~\ref{fig1}. Experimental results on FD-7 and FSD-10 \cite{a55}  demonstrate that our method achieves state-of-the-art performance. Our contributions can be summarized as follows.
\begin{itemize}
\item We propose a plug-and-play Cross-block Fine-grained Semantic Cascade (CFSC) Module capable of capturing ignored fine-grained information from GCN backbone.
\item We collected the \textit{FD-7 fencing dataset}, including seven classes of high-speed fencing actions. There are subtle distinctions between different categories.
\item Our approach significantly outperforms state-of-the-art methods on both self-built FD-7 and the public FSD-10 figure skating datasets.
\end{itemize}


\section{Related Work}


\subsection{Skeleton-based Action Recognition with GCNs}
Research on GCNs can be broadly categorized into two approaches: spectral-based methods \cite{a101, a102, a102} and spatial-based methods \cite{a104, a105}. Spectral-based methods rely on the Laplacian eigenbasis, which can be applied only to graphs with the same structure. Spatial-based methods extend convolution networks to non-Euclidean spaces. They directly perform convolution operations on the features of graph nodes and their neighboring nodes using the human body's topological structure graph. In this work, we adopt the spatial-based approach to research.

Skeleton-based action classification methods \cite{a40, a39, a38, a37, a36, a35, a34, a33, a32} effectively capture the intrinsic characteristics of motions and exhibit strong robustness. Skeleton-based methods can be broadly categorized into two types: handcrafted-based methods and deep learning-based methods. Handcrafted-based methods emphasize features designed based on physical intuition, such as distances and joint angles \cite{a201, a202}, which not only require considerable effort for manual feature engineering, but also fail to extract high-level latent features. With the advancement of deep learning, data-driven automatic feature learning methods have become mainstream \cite{xia2022incomplete, jing2023marginalized}. Yan et al. \cite{a23} propose a spatial-temporal graph convolution network that directly models skeleton data as a graph structure. To address the problem of predefined topological structures that do not capture long-range dependencies between joints, Li et al. \cite{a25} introduce an action interaction module to infer behavior links that capture action-specific latent dependencies, combining activity links with structural links to form a generalized skeleton graph. However, these operations primarily focus on enhancing model structures within the same GCN block and do not leverage the low-level fine-grained features from shallow blocks.

\subsection{Fine-Grained Classification}
\begin{figure}[h]
    \centerline{\includegraphics[width=80mm, height=35mm]{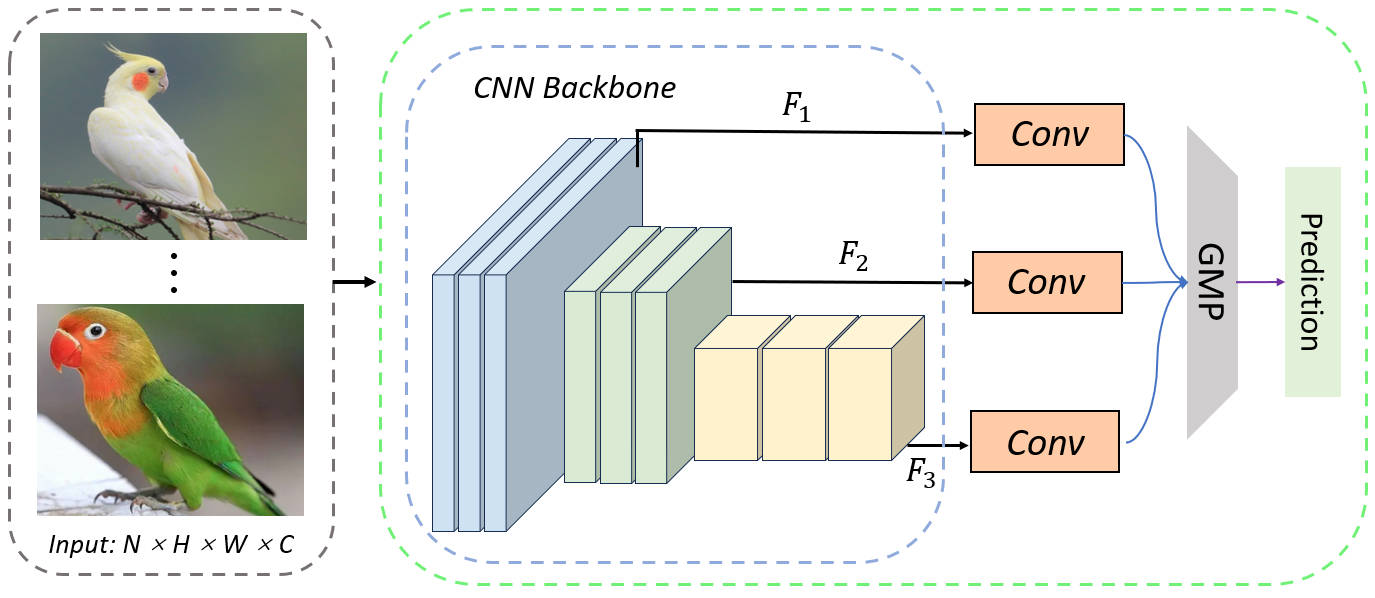}}
    \caption{Partial schematic diagram of the network in \cite{a041} where potential fine-grained information is captured by aggregating feature maps from different depth blocks.}
    \label{fig2}
\end{figure}

Fine-grained classification requires distinguishing subclasses within a certain parent class. Taking image classification as an example, conventional tasks aim to distinguish between cats and dogs, whereas fine-grained classification tasks require differentiation among specific breeds, such as ``Beagle,'' ``Husky'' within the "dog" category. The differences between subcategories are more subtle. Consequently, fine-grained classification tasks pose greater challenges.

Existing research \cite{a41, a42, a43, a44, a45, a46, a48, a49, xia2020hgnet} indicates that the shallow layers of the network contain rich low-level features that are crucial for fine-grained classification. As shown in Fig.~\ref{fig2}, Yang et al. \cite{a041} extract feature maps from different depth blocks, feed them into a global maximum pooling (GMP) layer, and then utilize aggregated features to generate prediction labels. In the context of sports, the postures and movements of the human body are generally similar, with only minor differences between the joints at certain moments. Thus, the classification of sports action also falls under the purview of fine-grained classification. However, existing action classification methods have not yet fully explored the features of different granularities at different levels. In this paper, we prove that fine-grained aggregated information from features at different levels enables superior performance in the classification of sports actions.

\begin{figure*}[t]
    \centerline{\includegraphics[width=16cm, height=7cm]{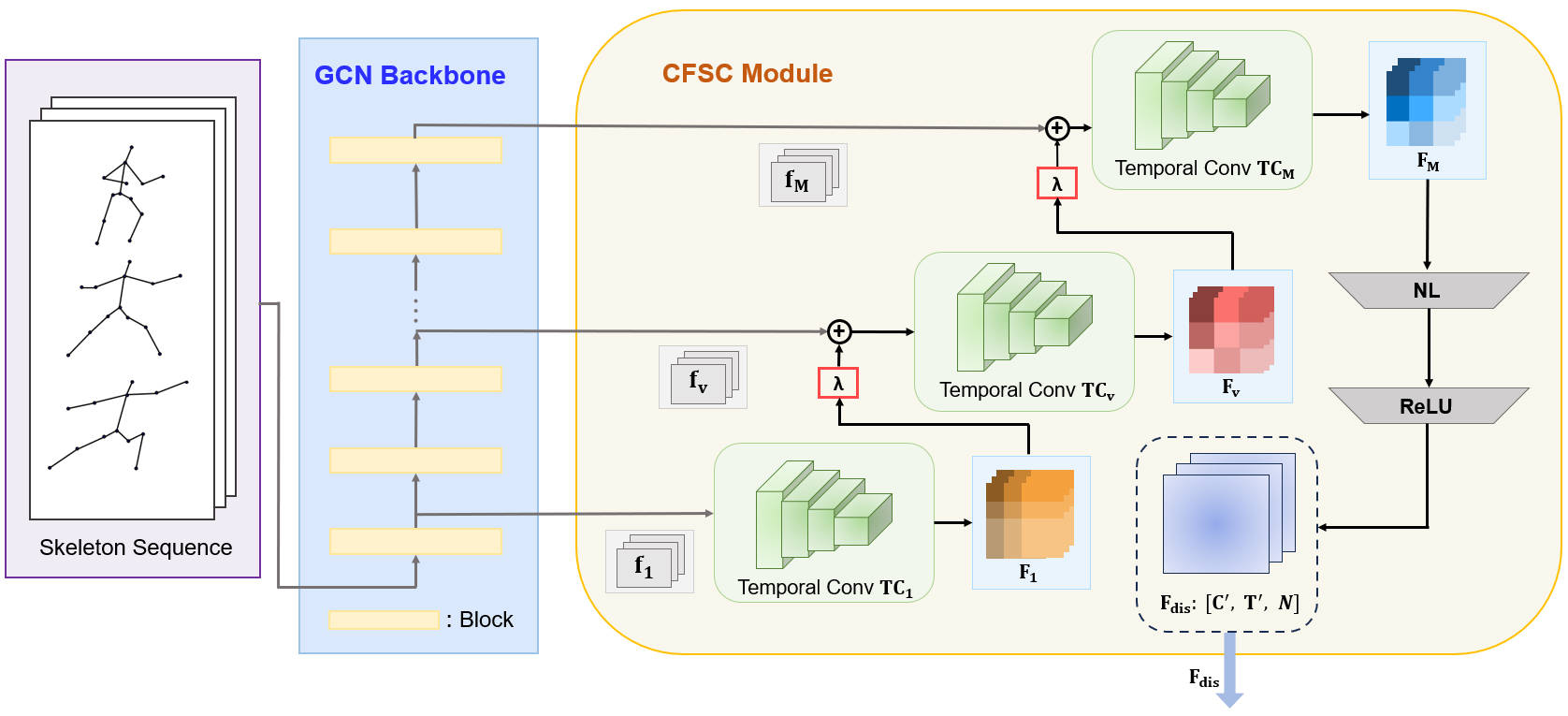}}
        \caption{Overview of CFSC Module. $M$ blocks of varying depths are selected from the GCN backbone to obtain a collection of feature maps that encompass different granularities of information: $\left\{\mathbf{f}_{1}, \mathbf{f}_{v},...,\mathbf{f}_{M}\right \}$. At each level, $\mathbf{f}_{v}$ is integrated with $\mathbf{F}_{v-1}$ obtained first from the preceding level. Next, temporal convolution is used to aggregate features of different granularities along the temporal dimension. The resulting feature $\mathbf{F}_{v}$ is then filtered by $\lambda$ and passed to the subsequent level. The aggregated highest-level feature $\mathbf{F}_{M}$, undergoes NL and ReLU, yielding the auxiliary discriminative feature $\mathbf{F}_{dis}$.}
    \label{fig3}
\end{figure*}
\section{Methodology}

In this section, we present a comprehensive description of the structure and underlying principles of the proposed CFSC module. First, STGCN \cite{a24} is exemplified as a reference to define relevant symbols and retrospectively review the basic structure of GCNs. Then, the operation mechanism of the CFSC is introduced. Moreover, we insert the CFSC Module into the backbone and introduce the overall architecture.


\subsection{Preliminaries}\label{sec3.1}
In the work of \cite{a24}, human skeleton sequences are used to construct an undirected spatial-temporal graph such as $\mathcal{G}=\left(\mathcal{V},\mathcal{E}\right)$. The node set $\mathcal{V}=\left\{{v}_{ti}|t=1,..., T,i=1,..., N\right\}$ represents the states of the $N$ nodes in each frame of a video with $T$ frames, and $\mathcal{E}=\left \{ {\mathcal{E}}_{\mathcal{S}},  {\mathcal{E}}_{\mathcal{T}} \right \}$ represents the edge sets used for spatial-temporal modeling.

For spatial dimension, the joints are connected according to the predefined human body topology to generate ${\mathcal{E}}_{\mathcal{S}}$, and the spatial adjacency matrix $\mathbf{A}\in \mathbb{R}^{N\times N}$ is used to represent the topological structure. In the graphs, the sampling function can be defined in the neighborhood set ${B}_{\mathcal{S}}\left ( {v}_{ti}\right )=\left \{ {v}_{tj}|d\left ( {v}_{tj},{v}_{ti}\right )\leq D\right \}$, where $D$ represents the distance between nodes ${v}_{tj}$ and ${v}_{ti}$. For the temporal domain, nodes ${v}_{ti}$ and ${v}_{\left (t+1\right)i}$ with the same index in consecutive frames are connected to generate ${\mathcal{E}}_{\mathcal{T}}$. The sampling function is defined as ${\mathcal{B}}_{\mathcal{T}}\left ( {v}_{ti}\right )= \left \{ {v}_{qi}| |q-t|\leq \left [ \mathcal{T}/2\right ]\right \}$. Eq. (\ref{eq1}) is used to implement ST-GCN in the spatial domain:
\begin{equation}
    \mathbf{f}_{out}=\displaystyle\sum_{k}^{{K}_{v}}\mathcal{W}_{k}\left ( \mathbf{f}_{in}\mathbf{A}_{k}\right)\odot \mathbf{M}_{k},
    \label{eq1}
\end{equation}
where ${K}_{v}$ corresponds to the kernel size of spatial dimension, $\mathcal{W}_{k}$ denotes the weighting function, $\mathbf{A}_{k}$ represents a normalized adjacency matrix, $\odot$ signifies the dot product operation and $\mathbf{M}_{k}$ refers to an attention map of size $N\times N$. For the temporal domain, the classical convolution operation ${K}_{t}\times 1$ is implemented to aggregate time characteristics.


\subsection{Structure of CFSC Module}\label{sec3.2}
Feature maps obtained from shallow blocks have been explored to collect
low-level information vital for the recognition of sports actions in \cite{a41, a42, a43}. However, they only used the output of features from the last block and suffered severe fine-grained information loss. To this end, the CFSC module is proposed to capture underlying fine-grained information in the backbone, thereby enhancing classification accuracy. 

The proposed module architecture is shown in Fig. \ref{fig3}. In summary, we extract features that contain rich low-level information from the GCN backbone and input them into the CFSC Module to generate auxiliary discriminative features, denoted as $\mathbf{F}_{dis}$. The input of the GCN backbone is a $C\times T\times N$ skeleton sequence that contains all motion information from the video clip. In the backbone, as the depth of the network increases, the information in the time dimension is continuously compressed, and the original low-order features are transformed into more complex high-order features. Consequently, the feature map gradually decreases along the time dimension, and the channel capacity gradually increases. The joint dimension remains unchanged, ensuring the continuous use of the human body topology map for the flow of information between joints. We select $M$ blocks, ranging from shallow to deep, and then extract and aggregate their fine-grained information. When selecting blocks, we adhere to the following three criteria. (1) Good coverage at varied depths of the network. The network is partitioned into three stages: shallow, medium, and deep, and features must be from at least two stages. (2) Discouraging features from adjacent blocks. The presence of comparable feature granularity within such blocks may lead to poor fusion performance. (3) Moderate number of blocks $M$. Fusing excessive features does not necessarily improve performance, but may significantly increase computational complexity.

In Sec. \ref{sec4.6}, the optimal blocks were empirically selected through experiments. $\left\{\mathbf{f}_{1}, \mathbf{f}_{v},...,\mathbf{f}_{M}\right \}$ in Fig. \ref{fig3} represent the feature maps output by $M$ selected blocks, where $\mathbf{f}_{1}$ represents the lowest-level feature, and $\mathbf{f}_{M}$ represents the highest ranking feature. We perform aggregation as follows:
\begin{equation}
    \mathbf{F}_{v}={TC}_{v}\left (\mathbf{f}_{v} \oplus \lambda \cdotp \mathbf{F}_{v-1} \right ),\label{eq2}
\end{equation}
where $\mathbf{f}_{v}$ represents the original feature produced by current block. The aggregation function $\oplus$ combines $\mathbf{f}_{v}$ and $\lambda \cdot \mathbf{F}_{v-1}$, where $\mathbf{F}_{v-1}$ is obtained by applying temporal convolution to the output $\mathbf{f}_{v-1}$ of the previous block and $\lambda$ is a scaling factor used to adjust the proportion. The aggregated features then undergo temporal convolution as ${TC}_{v}$, and the resulting $\mathbf{F}_{v}$ is utilized for the aggregation of the features at the next level. For the shallowest output feature $\mathbf{f}_{1}$, where no information is available from the previous level, a direct temporal convolution operation is performed. For the selected deepest feature $\mathbf{f}_{M}$, after undergoing temporal convolution, it is passed directly to the normalization layer.

\textbf{Temporal Convolution.} Each GCN block incorporates a temporal convolution module to aggregate features along the temporal dimension. However, as block operations are repeated, low-level features tend to be weakened, leading to significant fine-grained information loss. This paper aims to address this issue by conducting specialized time convolution operations on feature maps of varying granularity before engaging in the subsequent level of feature aggregation.

In the CFSC Module, the features of each level are input into the corresponding temporal convolution layer $\left \{ {TC}_{1},{TC}_{2},...,{TC}_{M}\right \}$, which serves two primary functions. (1) Aggregating temporal information by convolutions with a kernel size of $\left ({K}_{t},1\right )$. Sports activities involve higher speeds compared to other actions, resulting in a more prominent presence of short-term temporal features. These features often serve as crucial discriminators for similar sports actions. Hence, selecting a suitably sized temporal convolution kernel to capture short-term temporal features is crucial for a precise classification. Extensive experiments are conducted in Sec. \ref{sec4.5} to determine the optimal value for the parameter $K_t$, and the stride of convolution operation is set to $\left ( str, 1\right )$. Adjusting $str$, we temporally align the feature maps of adjacent hierarchical levels, ensuring compatibility and smooth information flow between different levels. (2) Unifying the number of channels. Different granularity features often have different dimensions. Through convolutional operations, feature maps of different granularities are aligned in the channel dimension to facilitate subsequent fusion operations.

\textbf{Feature Aggregation.} Five methods are explored for the aggregation of feature $\mathbf{f}_{v}$ and feature $\mathbf{F}_{v-1}$: Average, Max, Concat, Element-wise Multiplication, and Element-wise Addition. Average can smooth features, but when there are significant differences in feature maps, it may lead to information loss. The Max operation cannot preserve other detailed information apart from the maximum value, potentially leading to information loss. Although the Concat operation preserves all information from two feature maps, it increases the dimension of the features, thus increasing the complexity of computation. Element-wise multiplication enhances the influence of correlated features and effectively reduces feature dimensionality. However, it may overly emphasize the common information between the two features, while neglecting their differences. Compared to other methods, Element-wise Addition can preserve all the information of two feature maps and enhance the complementarity of features.
Hence, Element-wise Addition method is utilized for the aggregation of features.

\textbf{Feature Weighting.} Applying varying proportions to the aggregation of $\mathbf{f}_{v}$ and $\mathbf{F}_{v-1}$ results in distinct fine-grained feature interactions, which consequently influence the final classification results. Hence, we multiply $\mathbf{F}_{v-1}$ by the parameter $\lambda$ to adjust the proportion of low-level features during feature aggregation. 

After such manipulation, features of various granularities are captured and aggregated, and the temporal receptive field is continuously increased through temporal convolutions in the aggregation process. Therefore, the feature $\mathbf{F}_{M}$ contains rich long-range fine-grained temporal dependencies, which not only reduce the ambiguity between actions, but also provide global context, assisting the model to better capture spatial-temporal features. Subsequently, $\mathbf{F}_{M}$ is input into layers \textbf{NL} and \textbf{ReLU} to obtain auxiliary discriminative features $\mathbf{F}_{dis}$. \textbf{NL} represents the normalization layer. Within the \textbf{NL} layer, the average value $\mathbf{F}_{mean}$ and the standard deviation $\mathbf{F}_{std}$ are calculated for each channel of feature $\mathbf{F}_{M}$, treating the channels as individual units. And the data are normalized using the formula:
\begin{equation}
\mathbf{F}_{M}'=\frac{\mathbf{F}_{M}-\mathbf{F}_{mean}}{\mathbf{F}_{std}}.\label{eq3}
\end{equation}
This improves training stability and expedites model convergence. And, the normalized feature is fed into the \textbf{ReLU} function to enhance the model's ability for nonlinear fitting.

\subsection{Overall Architecture}\label{sec3.3}
\begin{figure}[h]
    \centerline{\includegraphics[width=85mm, height=85mm]{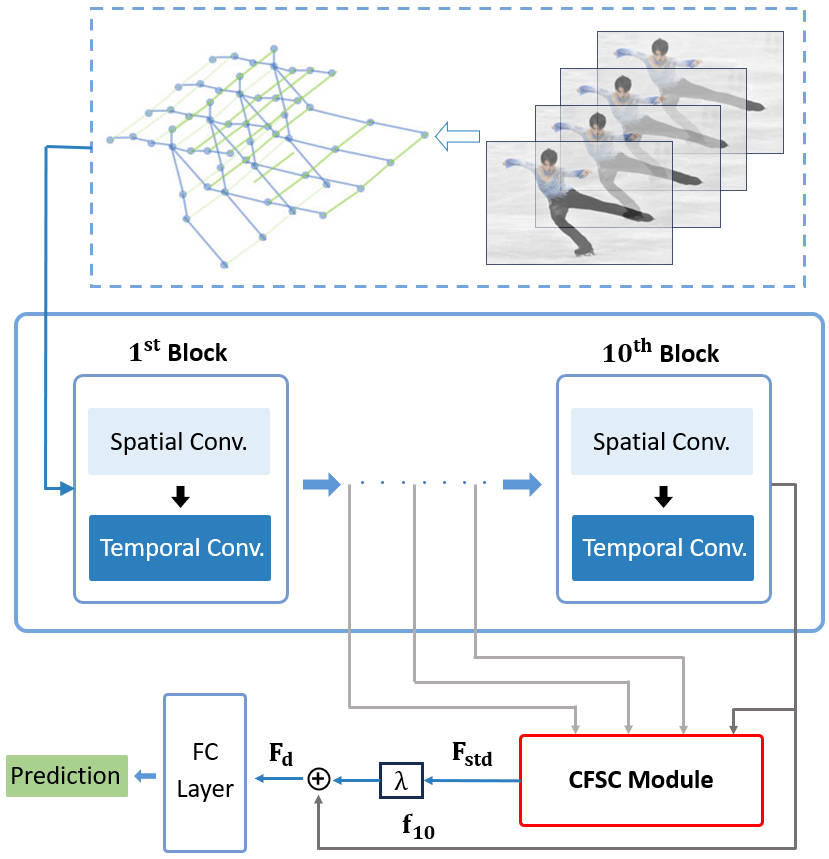}}
    \caption{Architecture overview. The HD-GCN backbone network, contained within the blue box, accepts skeletal sequences as input. The auxiliary feature $\mathbf{F}_{std}$ generated by the CFSC Module is aggregated with $\mathbf{f}_{10}$ and subsequently passed into the FC layer for generating prediction labels.}
    \label{fig4}
\end{figure}

As shown in Fig. \ref{fig4}, we adopt \cite{a38} as our baseline network architecture and embed our CFSC Module in it. 
The entire baseline network comprises 10 basic blocks. Within each block, spatial convolution operations are applied to aggregate spatial features, followed by temporal convolution operations to aggregate temporal features across frames. Residual connections are employed in each block to ensure training stability. The number of output feature channels for blocks 1 to 10 is 64-64-64-128-128-128-256-256.
At the 5-th and 8-th block, time dimension convolution stride is set to 2 to halve the temporal dimension. Based on the experimental results in Sec. \ref{sec4.6}, $M$ blocks are selected from 10 GCN blocks, and their outputs are used as input for the CFSC module. CFSC utilizes the multi-scale features provided by the baseline network to generate auxiliary discriminative features $\mathbf{F}_{std}$. Subsequently, our network uses the hyperparameter $\lambda$ to adjust the proportion of $\mathbf{F}_{std}$ in the feature aggregation process. Element-wise Addition operation is employed to aggregate the feature $\mathbf{F}_{std}$ with the output $\mathbf{f}_{10}$ obtained from the final block, producing the ultimate discriminative feature $\mathbf{F}_{d}$. Last, $\mathbf{F}_{d}$ is fed into a global average pooling layer and a softmax classifier to predict the sports action label.

\section{Experiments}

\begin{figure*}[ht]
    \centerline{\includegraphics[width=17cm, height=8cm]{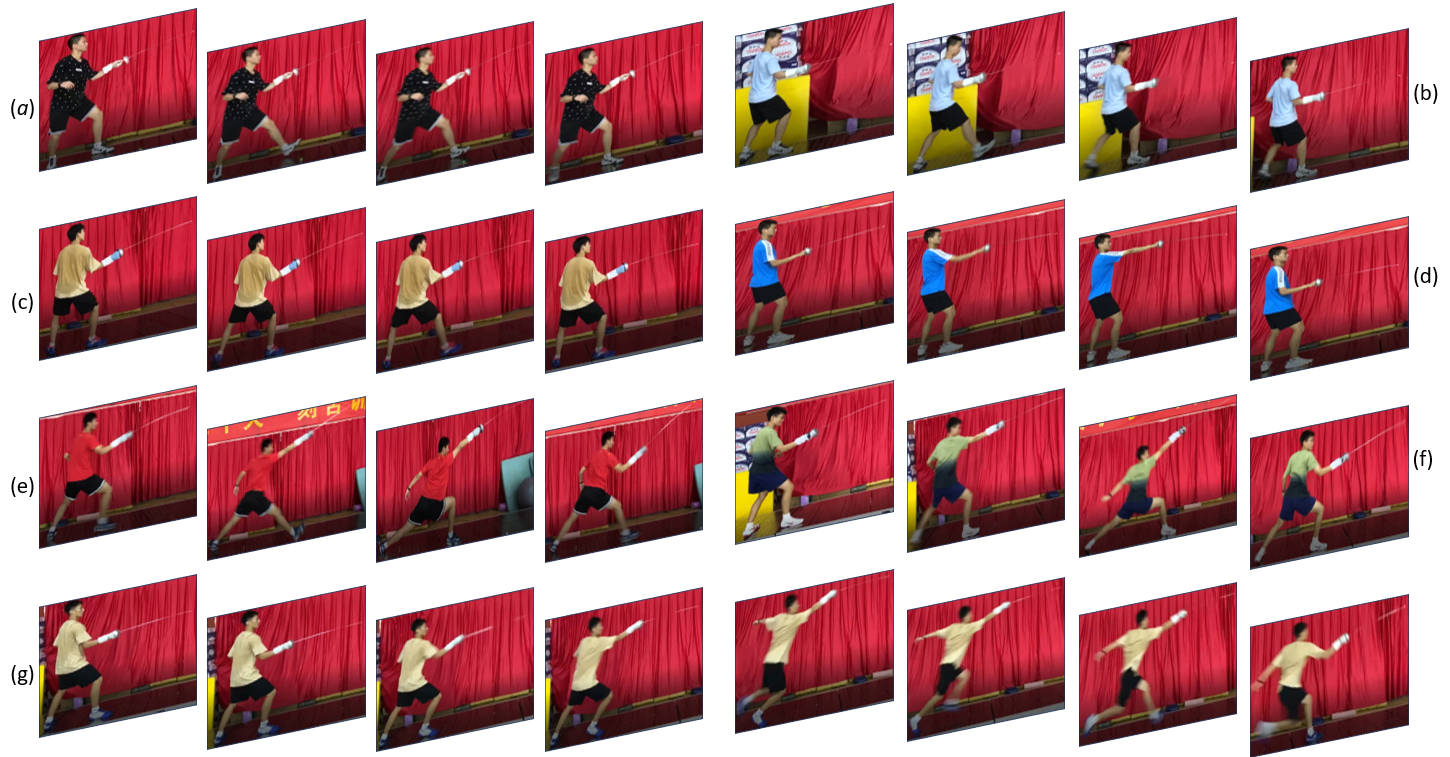}}
    \caption{Several examples from the FD-7 dataset are displayed in the figure. Action classification in the same sport involves precise categorization with slight differences between classes. For fencing footwork $\left (a\right )$, $\left (b\right )$, $\left (c\right )$, the fencer maintains a standard stance with the right foot forward and toes pointing straight ahead. During offensive actions $\left (e\right )$, $\left (f\right )$, $\left (g\right )$, the fencer leans forward with body weight shifted and executes an upward strike using the sword hand. The extraction and use of fine-grained information within the network is recommended to enhance sports action classification accuracy.}
    \label{fig5}
\end{figure*}
\begin{figure}[htb]
    \centerline{\includegraphics[width=50mm, height=40mm]{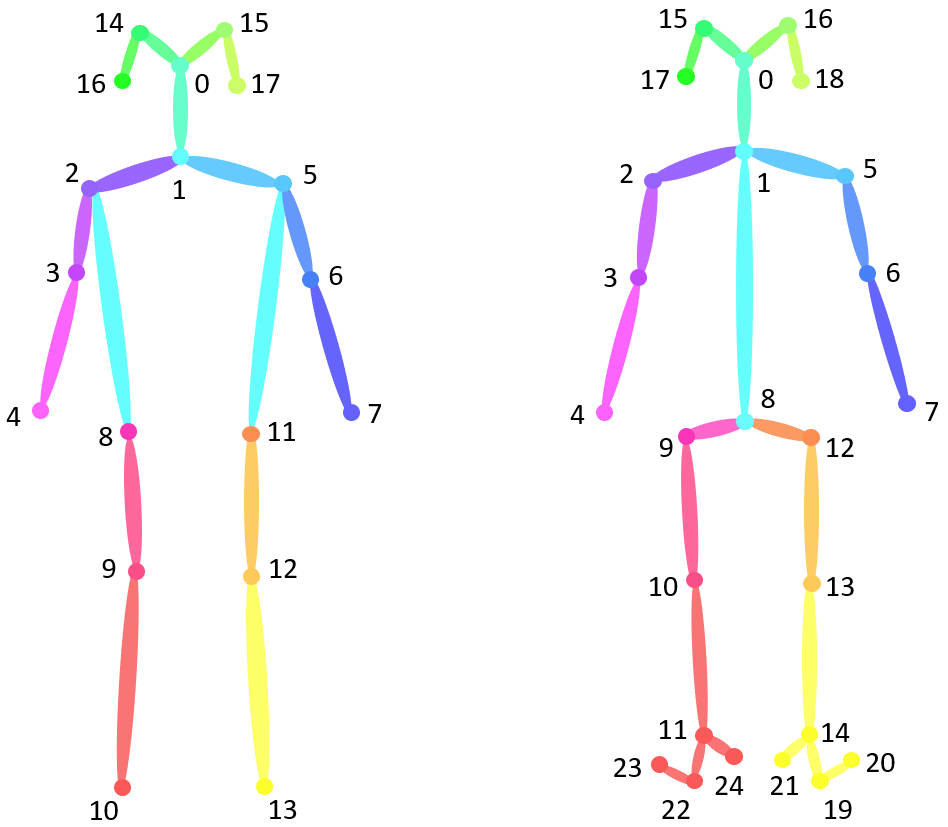}}
    \caption{The left sketch shows joint label of FD-7 dataset and the right sketch shows joint label of FSD-10 dataset.}
    \label{fig6}
\end{figure}

\subsection{Datasets}\label{sec4.1}
\textbf{FD-7.} The sport of fencing features swift sword velocities and a diverse repertoire of footwork actions, making it well suited for the fine-grained action classification task. To this end, we created a FD-7 fencing dataset, which was recorded with the help of 20 professional athletes from Nanjing Sport Institute, consisting of 10 females and 10 males, respectively. It ensures standardized actions and has substantial research value. As depicted in Fig. \ref{fig5}, FD-7 consists of seven categories: (a) step forward, (b) two-step forward, (c) step backward, (d) thrust in place, (e) lunge in place, (f) step forward lunge, and (g) sprint. These categories include regular fencing movement patterns and offensive actions. The dataset comprises 1,193 video clips, recorded at a frame rate of 30 frames per second and a camera resolution of $1920\times1080$. The length of each clip ranges from 1 to 5 seconds, representing a single action of one subject. The specific quantity of each action is presented in the Tab. \ref{table1}. When partitioning the dataset into training and validation sets, we ensured that the actions in both sets were executed by different athletes to prevent overfitting.

\begin{table}[h]
\centering
\renewcommand{\arraystretch}{1.2}
\begin{tabular}{cccc}
\hline
\textbf{Action Types} & \textbf{Training Set} & \textbf{Val Set} & \textbf{Total} \\
\hline
\hline
Step Forward       & 154          & 38      & 192  \\
Two-step Forward   & 150          & 38      & 188  \\
Step Backward      & 152          & 37      & 189  \\
\hline
Thrust in Place    & 81           & 33      & 114  \\
Lunge in Place     & 134          & 45      & 179  \\
Step Forward Lunge & 129          & 50      & 179  \\
\hline
Sprint             & 111          & 41      & 152  \\
\hline  
\end{tabular}
\caption{The number of clips for each category in FD-7.}
\label{table1}
\end{table}

\textbf{FSD-10.} Liu et al. \cite{a55} created a Figure Skating Dataset FSD-10 for fine-grained sports content analysis. FSD-10 comprises 1,484 videos, which encompass ten different categories of figure skating action in men / women programs, such as $2Axel$, $3Axel$, $3Loop$, $ChoreoSequence1$, $StepSequence3$, $FlyCamelSpin4$, $ChComboSpin4$, and others. All video materials were collected from the figure skating championships held between 2017 and 2018, and the duration of each video varies from 3 to 50 seconds.


FSD-10 has several appealing properties. First, actions in FSD-10 are characterized by higher speed and greater complexity. For instance, the complex 2Axel jump is completed in only about 2s. It is worth noting that the jump type heavily depends on the take-off process, which is a hard-captured moment. Second, actions of FSD-10 are original from figure skating competitions, which are consistent in type and sports environment (only skater and ice around). The above two aspects create difficulties for the machine learning model.



\subsection{Experimental Settings}\label{sec4.2} 
For FD-7, Openpose API \cite{a2} is utilized to convert original RGB videos into skeletal sequences of dimensions $N\times C\times T\times N$, where $C$ represents two-dimensional coordinates $\left ( x,\ y\right )$ and confidence scores $z$ of each keypoint, with a total of $N=18$ keypoints, and $T$ represents the length of the video. For simplicity, each clip is padded by replaying the sequence from the beginning to achieve $T = 150$. Also, the batch size is set to 16. FSD-10 also utilizes the Openpose API to capture 2D skeletal sequences. Due to the larger number of intricate movements involved in skating, $N$ is increased to 25 to capture a more complete set of information, and the parameter $T$ is set to 1,500. The joint labels for the FD-7 and FSD-10 datasets are showed in Fig. \ref{fig6}.

\begin{table*}[t]
\centering
\renewcommand{\arraystretch}{1.2}
\begin{tabular}{l|c|c|cc|cc}
\hline
                          &                               &                               & \multicolumn{2}{c|}{FD-7}                                                     & \multicolumn{2}{c}{FSD-10}                                                    \\
\multirow{-2}{*}{Methods} & \multirow{-2}{*}{Publication} & \multirow{-2}{*}{CFSC Module} & Top-1(\%): Joint                      & \multicolumn{1}{c|}{Top-1(\%): Bone}  & Top-1(\%): Joint                      & Top-1(\%): Bone                       \\
\hline
\hline
2S-AGCN \cite{a24}                  & CVPR 2019                     & \ding{55}                             & 61.8                                  & 51.1                                  & 52.9                                  & 80.7                                  \\
2S-AGCN \cite{a24}                  & CVPR 2019                     & \ding{51}                             & \textbf{78.2}                         & \textbf{63.6}                         & \textbf{53.2}                         & \textbf{84.0}                         \\
CTR-GCN \cite{a33}                 & ICCV 2021                     & \ding{55}                             & 79.3                                  & 52.1                                  & 85.7                                  & 87.1                                  \\
CTR-GCN \cite{a33}                 & ICCV 2021                     & \ding{51}                             & \textbf{91.1}                         & \textbf{57.5}                         & \textbf{88.0}                         & \textbf{89.4}                         \\
HD-GCN \cite{a38}                  & ICCV 2023                     & \ding{55}                             & 93.6                                  & 98.2                                  & 85.9                                  & 88.2                                  \\
HD-GCN \cite{a38}                  & ICCV 2023                     & \ding{51}                             & \cellcolor[HTML]{F8FF00}\textbf{95.7} & \cellcolor[HTML]{F8FF00}\textbf{99.6} & \cellcolor[HTML]{F8FF00}\textbf{88.2} & \cellcolor[HTML]{F8FF00}\textbf{90.1}                                           \\
\hline
\end{tabular}
\caption{Recognition performance of 2S-AGCN \cite{a24}, CTR-GCN \cite{a33}, and HD-GCN \cite{a38} models before and after integrating CFSC Module on FD-7 and FSD-10 datasets is shown in the table. The third column indicates whether a CFSC is inserted. The Top-1 accuracy is obtained using joint and bone input, with the best performance highlighted in yellow cells.}
\label{table2}
\end{table*}

HD-GCN \cite{a38} is adopted as our baseline. When conducting experiments using the baseline, SGD optimizer is employed with a Nesterov momentum of 0.9 and a weight decay of 0.0004. For the purpose of gradient back propagation, the cross-entropy loss function is selected. The training process encompasses 90 epochs, with an initial warm-up strategy \cite{a301} applied for the first five epochs to ensure a more stable learning process. The learning rate follows a cosine annealing schedule \cite{a302}, ranging from a maximum of 0.1 to a minimum of 0.0001. When experiments are conducted using other models, experimental details are strictly adhered to in accordance with the requirements specified in their paper. In the experiment, the recognition accuracy decreased when merging the outputs of different modalities, such as joint and bone. One possible explanation is that due to the fast pace and small inter-class differences in sports actions, when the input modality varies, the output features often exhibit significant disparities. Aggregating outputs from different modalities, on the contrary, may disrupt the representation of motion characteristics. Therefore, we decide not to consider the fusion of multiple input streams. All experiments were performed using a single RTX 3090 GPU.

\subsection{Comparison with State-of-the-Arts Methods}\label{sec4.3}

We insert the CFSC module into (1) 2S-AGCN \cite{a24}, (2) CTR-GCN \cite{a33}, (3) HD-GCN \cite{a38} networks, using joint and bone inputs, and evaluate the model accuracy on FD-7 fencing dataset and FSD-10 figure skating dataset. FD-7 dataset has $T=150$ frames in each video sequence. It represents sports activities with brief, subsecond durations. FSD-10 dataset, with $T=1,500$, represents sports activities characterized by prolonged durations measured in minutes. As shown in Tab. \ref{table2}, the state-of-the-art method HD-GCN obtains an action classification accuracy of 93.6\% and 98.2\% for the joint-input and bone-input on FD-7, 85.9\% and 88.2\% of classification accuracy for joint-input and bone-input on FSD-10. Upon incorporating CFSC Module into HD-GCN, notable improvements are achieved with an increase of 2.1\% and 1.4\% in classification accuracy for joint-input and bone-input on the FD-7 dataset. Similarly, on the FSD-10 dataset, the joint-input and bone-input classification accuracies experience enhancements of 2.3\% and 1.9\%, respectively.

Furthermore, after integrating our CFSC Module into the 2S-AGCN \cite{a24} and CTR-GCN \cite{a33} networks, significant performance improvements are observed across both datasets. The experimental results demonstrate that by extracting and aggregating features from shallow blocks, which have been overlooked in previous studies, fine-grained latent information of sports activities is effectively captured, leading to improved classification accuracy. After incorporating our CFSC Module into different models, performance improvements are observed on both the ``a few seconds actions'' and ``long-duration actions'' datasets, which indicates that our method demonstrates robust generalization capabilities.

It is worth noting that our work is the first to consider aggregating multi-granularity features from different blocks and applying them to GCN-based action classification models. In the upcoming sections, extensive experiments have been done on the selection of certain parameters in the model.

\subsection{Experiments on Parameter \textbf{$\lambda$}}\label{sec4.4}

\begin{figure}[ht]
    \centerline{\includegraphics[width=80mm, height=52mm]{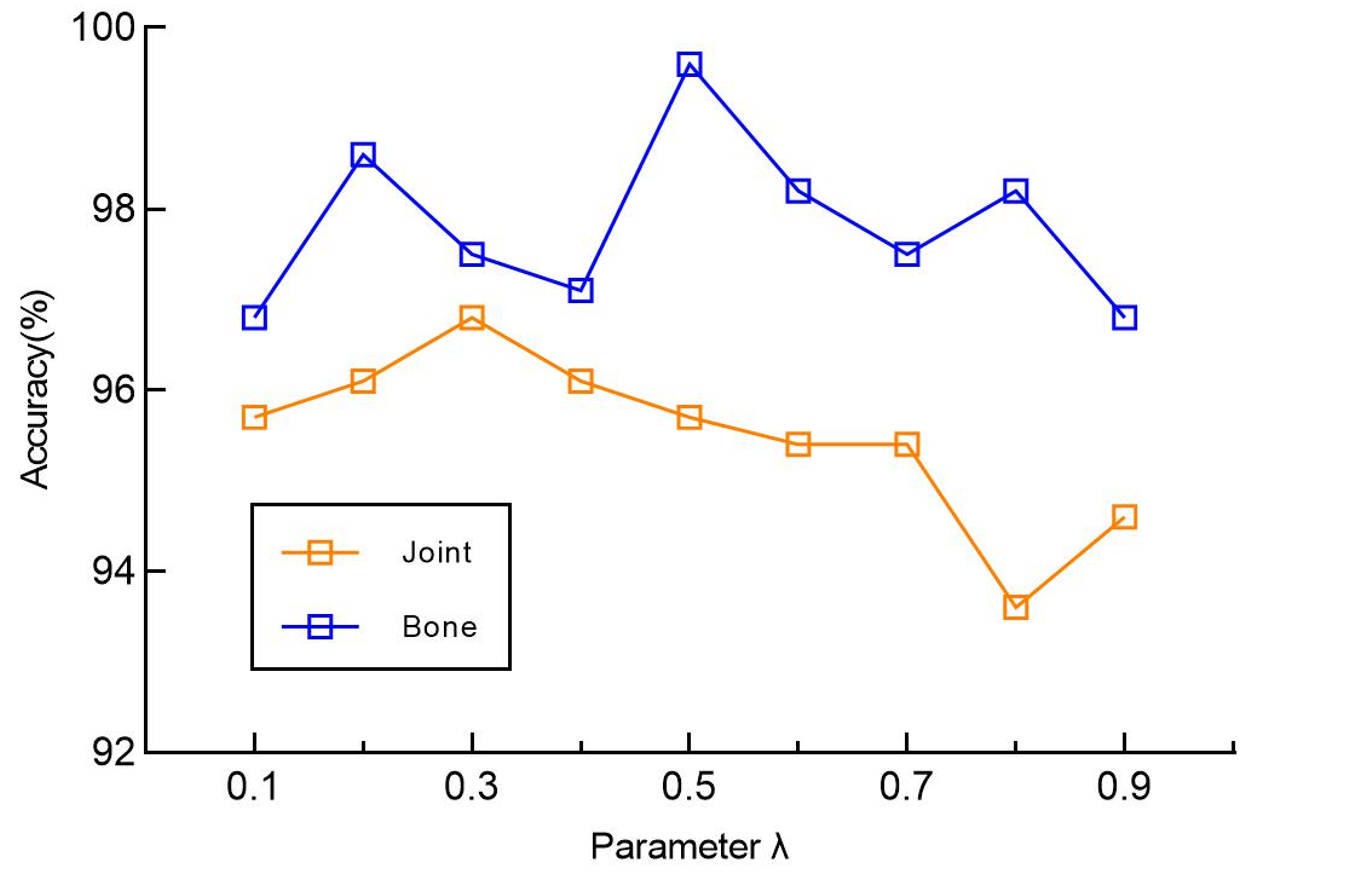}}
    \caption{Impacts of parameter $\lambda$ on the classification accuracy on the FD-7 datasets.}
    \label{fig7}
\end{figure}

In this section, the impact of the parameter $\lambda$ on classification accuracy is investigated using the FD-7 dataset. The value of $\lambda$ within the proposed CFSC Module determines the proportion of low-level features during feature aggregation. We set $\lambda \in \left [ 0.1, 0.9\right ]$ with a sampling interval of 0.1, resulting in 18 experimental groups. The experimental results are illustrated in Fig. \ref{fig7}.

Overall, as $\lambda$ increases, the accuracy of both input streams initially improves and then declines. For the joint stream, the peak is achieved at $\lambda$ = 0.3, surpassing the mentioned state-of-the-art method by 3.2\%. For the bone stream, the peak is achieved at $\lambda$ = 0.5, surpassing the mentioned state-of-the-art method by 1.4\%. These findings indicate that aggregating low-level features and high-level features in an appropriate proportion can enhance recognition accuracy. When $\lambda$ is too small, the utilization of low-level fine-grained features is insufficient, resulting in minimal performance improvement. Conversely, when it is too large, an excess of low-level fine-grained auxiliary features can disrupt the original high-level features and lead to a decrease in recognition accuracy.

Moreover, we attempt to train the model to automatically learn the optimal $\lambda$ parameter, and joint stream achieves 89.3\% accuracy, while bone stream achieves 94.6\% accuracy. This indicates that $\lambda$ cannot be automatically learned and fine-tuned by the model.

\begin{table}[ht]
\centering
\renewcommand{\arraystretch}{1.2}
\begin{tabular}{ccc}
\hline
Time Kernel Size & Top-1(\%): Joint & Top-1(\%): Bone \\
\hline
\hline
3                    & 95.7            & \textbf{99.6}           \\
5                    & 95.7            & 96.1           \\
7                    & \textbf{98.2}            & 97.9           \\
9                    & 96.1            & 96.8           \\
11                   & 94.6            & 96.4           \\
\hline
\end{tabular}
\caption{Comparisons of validation accuracy on FD-7 dataset with different temporal kernel sizes.}
\label{table3}
\end{table}

\begin{table}[ht]
\centering
\renewcommand{\arraystretch}{1.2}
\begin{tabular}{ccc}
\hline
Block Number & Top-1(\%):   Joint & Top-1(\%): Bone \\
\hline
\hline
1, 10        & \textbf{97.9}    & 97.5            \\
4, 10        & 95.4             & 96.8            \\
7, 10        & 96.1             & 98.9            \\
4, 7, 10     & 95.7             & \textbf{99.6}   \\
1, 5, 10     & 94.6             & 97.5            \\
1, 4, 7, 10  & 94.3             & 97.5           \\
\hline
\end{tabular}
\caption{Comparisons of validation accuracy on FD-7 dataset with different block numbers.}
\label{table4}\vspace{-3mm}
\end{table}

\begin{table}[h]
\centering
\renewcommand{\arraystretch}{1.2}
\begin{tabular}{c|c|cc|cc}
\hline
\multirow{2}{*}{Action} & \multirow{2}{*}{CFSC Module} & \multicolumn{2}{c|}{Foot}       & \multicolumn{2}{c}{Hand}        \\
                        &                              & Left           & Right          & Left           & Right          \\
\hline
\hline
b                     & \ding{55}                           & 0.046          & 0.048          & \textbf{0.072}         & 0.069          \\
b                     & \ding{51}                          & \textbf{0.084} & \textbf{0.054} & 0.051          & \textbf{0.076} \\
f                     & \ding{55}                           & 0.085          & \textbf{0.070}          & 0.096          & 0.099          \\
f                     & \ding{51}                          & \textbf{0.102} & 0.069          & \textbf{0.138} & \textbf{0.279}  \\
\hline
\end{tabular}
\caption{The numerical values of critical joints during the execution of actions (b) and (f), both prior to and subsequent to the insertion of the CFSC Module.}
\label{table5}\vspace{-3mm}
\end{table}

\subsection{Effect of Temporal Convolution Kernel Size}\label{sec4.5}
As depicted in Fig. \ref{fig3}, temporal convolution operations are employed on features of different granularities separately. In this section, extensive experiments are conducted on the size of temporal convolution kernel ${K}_{t}$ as illustrated in Tab. \ref{table3}. The results demonstrate small ${K}_{t}$ significantly enhance recognition accuracy: the joint stream reaches its peak at ${K}_{t}=7$, and the bone stream achieves its peak at ${K}_{t}=3$. 

However, as ${K}_{t}$ exceeds 7, the recognition accuracy starts to decline. This phenomenon is attributed to the fact that, compared to other actions, sports activities are characterized by fast motion and repetitive sub-actions, which require aggregation of short-term temporal features. The application of oversized temporal convolutions directly on lower-level features introduces unnecessary noise interference.

\subsection{Selecting Blocks}\label{sec4.6}

Blocks at different depths generate feature maps with diverse granularities of information. Tab. \ref{table4} illustrates the variations in recognition accuracy on FD-7 when different block output feature maps are used as inputs to CFSC Module. The joint stream achieves the highest accuracy when blocks 1 and 10 are selected. Similarly, skeletal stream achieves the highest accuracy when blocks 4, 7 and 10 are selected. Note that selecting an excessive number of blocks can lead to a decrease in recognition accuracy. One possible reason is that an excessive number of hierarchical features can introduce information redundancy. Our future work will focus on investigating self-learning methods to automatically determine the module selection scheme.

\subsection{Feature Visualization}\label{sec4.7}

\begin{figure}[ht]
    \centerline{\includegraphics[width=80mm, height=68mm]{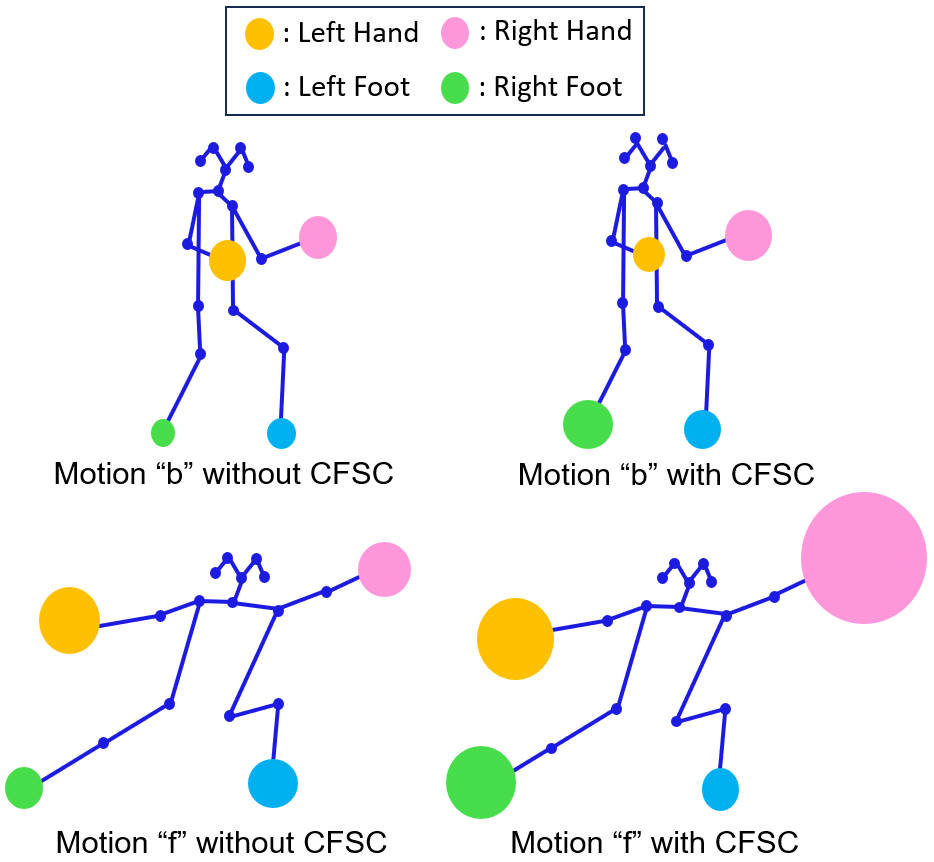}}
    \caption{Visualization of critical joints feature responses during the execution of actions (b) Two-step Forward and (f) Step Forward Lunge. The circular areas represent the magnitudes of the response.}
    \label{fig8}\vspace{-2mm}
\end{figure}

For fencing actions, the holding hand and the feet are critical joints for classification. In Fig. \ref{fig8}, we visualize the response of the left and right hands, as well as the feet, during action (b) Two-step Forward and action (f) Step Forward Lunge. Before incorporating the CFSC Module, feature map $\mathbf{f}_{10}$, obtained from the last block of the original backbone, is utilized to compute the feature response. After integrating the CFSC Module, feature $\mathbf{F}_{dis}$ is used to calculate the response to features of the critical joints.

Tab. \ref{table5} shows the specific response values. In Fig. \ref{fig8}, each circle surrounding a joint represents the magnitude of its feature response. After incorporating CFSC Module, feature response of the feet is enhanced in action (b), while the feature response of the holding hand is significantly improved in both actions (b) and (f). These results demonstrate that the CFSC Module is capable of enhancing the feature response of critical nodes, thus improving the accuracy of action recognition.



\section{Conclusion}
In this work, we proposed a plug-and-play CFSC method for skeleton-based sports action recognition. Feature maps generated by different depth blocks of GCN network were utilized as input to CFSC Module. Temporal information of different granularity features was extracted and aggregated from shallow to deep levels to generate auxiliary discriminative features. This approach effectively captured fine-grained features and improved the accuracy of sports action classification. Our method was evaluated on self-collected fencing dataset FD-7 and the publicly available dataset FSD-10. Compared to existing baselines, CFSC has significant advantages in improving performance in most cases.




{\small
\bibliographystyle{ieee}
\bibliography{egbib}
}

\end{document}